\documentclass{article}

\usepackage[english]{babel}
\usepackage[letterpaper,top=2cm,bottom=2cm,left=3cm,right=3cm]{geometry}
\usepackage{amsmath}
\usepackage{amssymb}
\usepackage{graphicx}
\usepackage{booktabs}
\usepackage{authblk}
\usepackage{cite}
\usepackage{verbatim}
\usepackage{embedfile}
\usepackage[colorlinks=true, allcolors=blue]{hyperref}

\input{glyphtounicode}
\embedfile[desc={BibTeX entry for this paper}]{forgione2027manifold.bib}

\newcommand{\base}{{base architecture}}
\newcommand{\ii}{{(i)}}
\newcommand{\tr}{\mathrm{tr}}
\newcommand{\te}{\mathrm{te}}
\newcommand{\norm}[1]{\left\lVert#1\right\rVert}
\newcommand{\seq}[1]{{\mathbf{#1}}}
\newcommand{\R}{\mathbb{R}}
\newcommand{\N}{\mathbb{N}}
\newcommand{\E}{\mathbb{E}}
\newcommand{\DD}{\mathcal{D}}
\newcommand{\loss}{\mathcal{L}}
\newcommand{\latent}{z}
\newcommand{\fullpar}{\theta}
\newcommand{\redpar}{\phi}
\newcommand{\projpar}{\gamma}
\newcommand{\encpar}{\psi}
\newcommand{\M}{F}
\newcommand{\CFD}{G}

\newtheorem{remark}{Remark}
\newtheorem{problem}{Problem}
\newtheorem{theorem}{Theorem}

\title{Manifold meta-learning for reduced-complexity\\neural system identification}

\author[1]{Marco Forgione}
\author[2]{Ankush Chakrabarty}
\author[1]{Dario Piga}
\author[1]{Matteo Rufolo}
\author[3]{Alberto Bemporad}
\affil[1]{SUPSI, IDSIA - Dalle Molle Institute for Artificial Intelligence, Lugano, Switzerland.}
\affil[2]{Mitsubishi Electric Research Laboratories (MERL), Cambridge, MA, USA.}
\affil[3]{IMT School for Advanced Studies Lucca, Lucca, Italy.}

\begin{document}

\maketitle

\begin{abstract}
System identification has greatly benefited from deep learning techniques, particularly for modeling complex, nonlinear dynamical systems with partially unknown physics where traditional approaches may not be feasible. However, deep learning models often require large datasets and significant computational resources for both training and inference due to their high-dimensional parameterizations.
To address this challenge, we propose a meta-learning framework that discovers a low-dimensional manifold within the parameter space of an over-parameterized neural network architecture. This manifold is learned from an abundant meta-dataset of input-output sequences generated by a simulator of the class of systems of interest, and thereafter enables efficient model training while preserving the network's expressive power for the considered system class. Unlike bilevel meta-learning approaches, our method employs an auxiliary neural network to map datasets directly onto the learned manifold, eliminating the need for costly second-order gradient computations during meta-training and generally reducing the number of first-order updates required in inference, which could be expensive for large models. Finally, we derive a probabilistic bound that quantifies the performance gap between training on the learned low-dimensional manifold and the full parameterization. We validate our approach on a family of Bouc--Wen oscillators, which is a well-studied nonlinear system identification benchmark. We demonstrate that we are able to learn accurate models even in small-data scenarios.
\end{abstract}

\noindent\rule{\textwidth}{1pt}
To cite this work, please refer to the published version:\\ \\
\textsf{Forgione, M., Chakrabarty, A., Piga, D., Rufolo, M., \& Bemporad, A. (2027). Manifold meta-learning for reduced\nobreakdash-complexity
neural system identification.} \textit{Automatica}, \textit{196}, Article 113333. \url{https://doi.org/10.1016/j.automatica.2026.113333}

\vskip 1.5em
\noindent BibTeX entry:
{\footnotesize
\begin{verbatim}
@article{forgione2027manifold,
title = {Manifold meta-learning for reduced-complexity neural system identification},
journal = {Automatica},
volume = {196},
pages = {113333},
year = {2027},
issn = {0005-1098},
doi = {https://doi.org/10.1016/j.automatica.2026.113333},
url = {https://www.sciencedirect.com/science/article/pii/S0005109826005145},
author = {Marco Forgione and Ankush Chakrabarty and Dario Piga and Matteo Rufolo and Alberto Bemporad}
}
\end{verbatim}}

\noindent\rule{\textwidth}{1pt}

\section{Introduction}
In recent years, system identification has advanced significantly with the adoption of deep learning tools~\cite{pillonetto2025deep, deshpande2023physics, Bem25, champneys2024baseline}. These methods have proven  effective in modeling complex, nonlinear dynamical systems where the underlying physics is partially or entirely unknown, relying on flexible and expressive black-box components such as neural networks. However, a fundamental challenge of black-box system identification is the need for large datasets to ensure model reliability across a wide range of operating conditions. Even when such datasets are available, training deep models with numerous parameters can be computationally demanding, limiting their practical applicability. This contrasts with physics-based modeling, which, when feasible, enables data-efficient learning and typically results in models that generalize well beyond the training regime.

To bridge the gap between black-box and physics-based approaches, recent research has explored the opportunity to include domain knowledge into deep-learning architectures and training algorithms, while preserving their flexibility. These efforts include designing custom model structures that satisfy known physical constraints by design~\cite{moradi2023physics, deshpande2023physics, forgione2021continuous} and introducing regularization terms in the loss function to promote the satisfaction of known dynamics, as seen in physics-informed machine learning~\cite{nghiem2023physics}. While highly successful, these approaches still require expert knowledge and manual tuning for the design of appropriate architectures and regularizers.

This paper addresses the challenge of learning simplified model architectures in an automated manner using meta-learning techniques~\cite{hospedales2021meta}. In line with the meta-learning setting for system identification~\cite{forgione2023system,chakrabarty2025meta},
we assume access to a \emph{meta-dataset}, namely, a collection of datasets generated by different, yet related, dynamical systems representative of a target population that we aim to describe. The proposed method requires an abundant meta-data regime: in practice, the meta-dataset is generated by a simulator of the class of
systems of interest, with parameters randomized within physically plausible ranges.

Additionally, we assume that a given black-box architecture, referred to as the \emph{\base}, is sufficiently expressive to capture the dynamics of each system in the meta-dataset for at least one (system-dependent) choice of its tunable parameters. However, this \base\ is highly over-parameterized, meaning it has significantly more degrees of freedom than the actual factors governing variability across systems in the meta-dataset.
Given the meta-dataset and the \base,\ our goal is to meta-learn a \emph{low-dimensional manifold} embedded in the \base's parameter space such that the \base,\ with parameters restricted to this manifold, can still describe all the systems within the class of interest.
Restricting the search space to this manifold effectively reduces the degrees of freedom  of the {\base}, thereby enhancing generalization of the estimated models.

Conceptually, our approach is inspired by   classic approaches for {model-based} meta-learning, which aim at \emph{optimizing the algorithm}  that estimates a model within a given \base~\cite{hospedales2021meta}. For instance,  model-agnostic meta-learning (MAML)~\cite{finn2017model} seeks an \emph{initialization} of the \base's\ parameters such that, with a few steps of gradient descent starting from that point, effective adaptation to each dataset in the meta-dataset is achieved. This meta-learning problem is tackled through bilevel optimization, where the initial parameters are optimized at the outer level, while adaptation to the specific datasets is performed via gradient descent at the inner level. Recently, the MAML framework has been adopted for the design of adaptive model-based control schemes in \cite{spencer2023control}.
Unlike MAML, our methodology meta-learns a low-dimensional manifold in the \base's\ parameter space instead of providing an initialization for optimization, and can be related to learning an effective {\it inductive bias} to facilitate the system identification procedure.

An existing meta-learning algorithm that even more closely resembles our approach is latent embedding optimization (LEO)~\cite{rusu2018meta}, which meta-learns a low-dimensional embedding of the datasets. Compared to LEO (and MAML), we circumvent the need to perform bilevel optimization by introducing an \emph{auxiliary neural network} that directly maps each dataset onto its low-dimensional representation, bypassing the need for gradient-based inner-loop optimization.

Meta-learning has been recently applied for system identification, among other works, in~\cite{du2023can,forgione2023system,chakrabarty2025meta}. However, \cite{du2023can,forgione2023system} build a unified \emph{meta-model} for the entire meta-dataset, relying on the \emph{in-context} learning capabilities of transformer architectures. Conversely,~\cite{chakrabarty2025meta} is based on the MAML algorithm and its variants; unlike the current approach, MAML-like approaches involve a bilevel training procedure. While this is effective, a major drawback is the complexity of the bilevel training procedure. To avoid this, workarounds such as implicit MAML (iMAML) have been investigated by~\cite{yan2024mpc}, which leverages the implicit function theorem to simplify the inner-loop training. Also,~\cite{chakrabarty2025meta} demonstrates the potential of almost-no-inner-loop (ANIL), which restricts the inner-loop training to a small set of user-chosen layers or components in the network.  In other words, ANIL restricts model adaptation to a  linear subspace, namely the span of the weights in the layers that are chosen to remain learnable.
In contrast, our method specifies only the dimensionality of the adaptation manifold, while allowing its structure to be meta-learned from data. As a result, our approach can recover ANIL as a very special case, but is not limited to it. Indeed, it can also discover more general manifolds, potentially involving non-linear parameter combinations rather than a manually predefined subset of layers.

\section{Framework}
\label{sec:framework}
\subsection{Data stream and distribution}
In line with the meta-modeling framework introduced in~\cite{forgione2023system}, we assume to have access to a large (possibly unlimited) collection $\DD$ of datasets referred to as the meta-dataset, described by
\begin{equation}
\label{eq:dataset_stream}
\DD=\{D^\ii\}_{i\in\N}, \text{ where } D^\ii = (\seq u_{0:N-1}^\ii, \seq y_{0:N-1}^\ii).
\end{equation}
Each dataset $D^\ii$ comprises $N$
inputs $\seq u^\ii_k \in \R^{n_u}$
and corresponding targets $\seq y^\ii_k \in \R^{n_y}, k=0,1,\dots,N-1$.
In the context of system identification, $\seq y^\ii_{0:N-1}$ is the sequence of outputs generated by a dynamical system $S^\ii$ driven by the input sequence  $\seq u^\ii_{0:N-1}$. This results in a causal  dependency $\seq u_{0:k}^\ii \rightarrow \seq y_k^\ii$. In the case of static regression, the dependency is simply $\seq u_k^\ii \rightarrow \seq y_k^\ii$.

In the following, we adopt the shorthand notation $\seq u=\seq u_{0:N-1}, \seq y = \seq y_{0:N-1}$ to denote the sequence of $N$ input and output samples, respectively, contained in a dataset.
The inputs $\seq u^\ii$ of each dataset are assumed to be drawn from a distribution with density $p(\seq u)$, while the dependency
$\seq u^\ii \rightarrow \seq y^\ii$
is different, yet related across datasets.
Formally, the collection of datasets~\eqref{eq:dataset_stream} may be interpreted as samples drawn from a distribution (in the space of datasets) with density:
\begin{equation}
\label{eq:dataset_distribution}
p(D) = p(\seq{u}, \seq{y}) = p(\seq{u})
\int_\latent p(\latent)
p(\seq{y} \mid \seq{u}, \latent) \; d\latent,
\end{equation}
where $\latent \in \R^{n_\latent}$ is an unknown or hidden variable that characterizes the data-generating mechanism $p(\seq{y} \mid \seq{u}, \latent)$. The  variable $\latent$  should be interpreted as an abstract representation of the factors of variation underlying the
data-generating process across datasets.
In the considered synthetic data generation setting, $\latent$ may correspond to a subset of the simulator's physical parameters, intentionally randomized to capture plausible variations of the input/output system dependency expected in the real-world scenarios of interest, such as
different masses attached to the end-effector of a robot, environment temperature in a chemical reaction,  etc.

\begin{remark}\label{rem:latent}
The latent variable $\latent$ only captures the aspects of the data-generating system that vary across datasets.
For instance, in the case of the robot where only the end effector mass changes, a scalar $\latent$ representing such mass suffices, even though the robot itself is characterized by many other physical parameters, that, however, remain fixed for all simulations. Those non-varying parameters may be thought of as part of the functional form of the conditional distribution $p(\seq{y} \mid \seq{u}, \latent)$ in~\eqref{eq:dataset_distribution}.
\end{remark}

\subsection{Base architecture}
\label{sec:full_order}
Consider the \base\ given by the black-box model structure
\begin{equation}
\label{eq:fullnet}
\hat {\seq{y}} = \M(\seq{u}; \fullpar),\;  \fullpar \in \R^{n_\fullpar},
\end{equation}
which we assume is prescribed to describe the input/output dependency $\seq u \rightarrow \seq y$ for each dataset $D$ in the support of $p(D)$ with practically no bias, for some system-dependent (unknown) value of the parameter $\fullpar$.
We argue that this assumption is not particularly restrictive given the availability of expressive function approximators such as deep neural networks, that, with appropriate selection of hyperparameters, can describe wide classes of nonlinear functions with high accuracy, and have been reported to exhibit competitive performance across different system identification benchmarks~\cite{champneys2024baseline}.

The model structure~\eqref{eq:fullnet} is typically characterized by a number of parameters $n_\fullpar \gg n_\latent$ for two reasons. First, \eqref{eq:fullnet} has a black-box structure, not necessarily tailored to a specific class of system. For example, accurate description of robot dynamics with $l=8$ links may require thousands of black-box parameters ($n_\fullpar \approx 10^3$) with neural-network models~\cite{Bem25}, whereas a physics-based representation only requires $10l = 80$ terms~\cite{gaz2019dynamic}.
Second, just a subset of these physical parameters may be varying across the simulations within the meta-dataset. For instance, as in the example of Remark~\ref{rem:latent}, if the only physical parameter varying across datasets is the robot's end-effector mass, then $n_\latent=1$ suffices.
Consequently, black-box architectures used to describe physical systems are often heavily over-parameterized.

For a given dataset $D=(\seq{u},\seq{y})$, the optimal parameter that best describes the input/output relation is typically obtained by minimizing a supervised learning criterion:
\begin{equation}
    \label{eq:full_sysid}
    \hat \fullpar = \arg \min_{\fullpar} \loss\left(\seq y, \hat{\seq y}(\fullpar)\right),
\end{equation}
where the training loss $\loss(\cdot, \cdot)$ penalizes the mismatch between measurements $\seq y$ and predictions $\hat{\seq y}(\fullpar)=\M(\seq u; \fullpar)$.
A common choice for $\loss(\cdot, \cdot)$
is the mean squared error $
\mathrm{MSE}(\seq y, \hat {\seq y}) = \frac{1}{N}\sum_{k=0}^{N-1}
\norm{\seq{y}_k -  \hat {\seq{y}}_k}_2^2$. A regularization term $r(\fullpar)$ may be added to~\eqref{eq:full_sysid}. A typical choice is $r(\fullpar)=\rho\|\fullpar\|_2^2$, where $\rho\geq 0$ controls the regularization strength.

However, when the number of parameters $n_\fullpar$ and/or the number of samples $N$ in the training dataset $D$ are large, learning the dynamics represented by $\M$ can be computationally intensive. Conversely, when $N$ is small, achieving good {\it generalization} to unseen data becomes difficult.

\subsection{Meta-learning manifolds in the parameter space}
To address the computational and generalization challenges of system identification within the base architecture~\eqref{eq:fullnet}, we consider the following problem:
\begin{problem}
Given  the base architecture~\eqref{eq:fullnet} and a collection $\DD$ of datasets sampled from $p(D)$,
learn a lifting function $P: \R^{n_\redpar} \mapsto \R^{n_\fullpar}$, with $n_\redpar \ll n_\fullpar$ that defines a $n_\redpar$-dimensional manifold $M=\{\fullpar\in\R^{n_\fullpar}$: $\fullpar=P(\redpar),\ \redpar \in \R^{n_\redpar}\}$
such that each dataset $D=(\seq{u},\seq{y})$ in the support of $p(D)$ can be described accurately by a model $\hat{\seq{y}}=\M(\seq{u};P(\redpar))$ for some coordinate vector $\redpar\in\R^{n_\redpar}$.
\label{prob:manifold}
\end{problem}
We label Problem~\ref{prob:manifold} as a \emph{meta-learning} problem, as our goal is to learn a unique manifold $M$ for the {\it entire meta-dataset}. Once the manifold is learned, the user can fit a model to any dataset $D=(\seq{u},\seq{y})$ contained in the support of $p(D)$ by solving a lower-dimensional system identification problem with
$n_\redpar \ll n_\fullpar$ optimization variables.

Problem~\ref{prob:manifold} admits at least one solution for $n_\redpar \geq n_\latent$. Indeed, we
hypothesize in Section~\ref{sec:full_order} that each data-generating system in the distribution (equivalently, each $\latent \in \R^{n_\latent}$) is represented by (at least) a $\fullpar \in \R^{n_\fullpar}$. Then, there exists a function $P^\dagger : \R^{n_\latent} \mapsto \R^{n_\fullpar}$ mapping each data-generating system (characterized by a $\latent \in \R^{n_\latent}$) to a parameter $\fullpar \in \R^{n_\fullpar}$ of the full black-box model structure.\footnote{The relationship $\latent \rightarrow \fullpar$ might be one-to-many. The function $P^\dagger$ here arbitrarily chooses one of the multiple equivalent solutions.}
Therefore, the model structure
$\hat {\seq{y}} = \M(\seq{u}; P^\dagger(\latent)),\;  \latent \in \R^{n_\latent}$, solves Problem~\ref{prob:manifold}
with $P=P^\dagger$, $\redpar=\latent$. However, the function $P^\dagger$, as well as $n_\latent$, are not generally known a priori.
Moreover, there is no guarantee that $P^\dagger$ is sufficiently regular to be learnable from data with practical numerical algorithms.

Notwithstanding this potential limitation, and therefore accepting that the resulting solution may be suboptimal, we set the dimensionality $n_\redpar \ll n_\fullpar$ of the reduced model parameter space
and describe the function $P$ by a vector $\projpar \in \R^{n_\projpar}$ of tunable parameters, denoting it by $P_\projpar$.
We assume that each function $P_\projpar$ in the \emph{user-chosen} search space $\{P_\projpar : \projpar \in \R^{n_\projpar}\}$ induces a differentiable manifold of dimension $n_\redpar$ embedded in $\R^{n_\fullpar}$.
For instance, $P_\projpar$ could be parameterized as an affine function or as a neural network with smooth activation functions.

As we will detail in Section~\ref{sec:reduced_order}, our meta-learning goal is to learn a lifting parameter vector $\hat \projpar \in \R^{n_\projpar}$ for the entire set $\DD$ to define the reduced-complexity architecture
\begin{equation}
\label{eq:rednet}
\hat {\seq{y}} = \M(\seq{u}; P_{\hat \projpar}(\redpar)),\;  \redpar \in  \R^{n_\redpar},
\end{equation}
given by the functional composition of the \base\ with  the \emph{learned} lifting function
$P_{\hat \projpar} : \R^{n_\redpar} \mapsto \R^{n_\fullpar}$.

Based on the learned reduced-complexity architecture~\eqref{eq:rednet}, and given a new dataset $D^* = (\seq{u}^*, \seq{y}^*)$ that is contained in the support of $p(D)$, the following reduced-complexity optimization problem
\begin{equation}
    \label{eq:reduced_sysid}
    \hat \redpar = \arg \min_{\redpar} \loss(\seq y^*, \M(\seq u^*; P_{\hat\projpar}(\redpar)))
\end{equation}
solves the identification problem of fitting a model to $D^*$.
Given the reduced number $n_\redpar$ of optimization variables, problem~\eqref{eq:reduced_sysid} can generally be solved more efficiently than~\eqref{eq:full_sysid}, e.g., by taking advantage of quasi-Newton~\cite{Bem25,Bem23} and even global optimization approaches that may not be computationally feasible in the full-complexity structure~\eqref{eq:fullnet}.
Furthermore, the risk of overfitting is mitigated by the lower number of degrees of freedom of the architecture~\eqref{eq:rednet} with respect to~\eqref{eq:fullnet}.

\section{Meta-learning procedure}
\label{sec:reduced_order}
In line with the standard meta-learning setting, we split each dataset
$D^\ii$ into training and test portions, i.e.,
\begin{equation}
\label{eq:dataset_stream_split}
D^\ii = (\seq{u}^\ii_\tr , \seq{y}^\ii_\tr ), (\seq{u}^\ii_\te , \seq{y}^\ii_\te ),\, i=1,2,\dots,
\end{equation}
where both $\seq{y}^\ii_\tr$ and $\seq{y}^\ii_\te$ are generated by the \emph{same} system $S^\ii$ for different input signals $\seq{u}^\ii_\tr$ and $\seq{u}^\ii_\te$, respectively.
Such a data split
allows one to learn an optimal manifold $M$ such that, when training is performed on $(\seq{u}^\ii_\tr , \seq{y}^\ii_\tr )$, performance measured on test data
$(\seq{u}^\ii_\te , \seq{y}^\ii_\te )$ is good.

A direct way of learning an optimal parametric function $P_{\hat \projpar}$ is to solve the following bilevel optimization problem:
\begin{subequations}
\label{eq:bilevel}
\begin{align}
\hat \projpar &= \arg \min_{\projpar} \E_{p(D)}
\big [ \loss({\seq y}_\te, \hat {\seq y}_\te) \big ] \label{eq:outer}\\
\hat{\seq{y}}_\te  &= \M\left(
  \seq{u}_\te ; \,
  P_{\projpar} ( \hat \redpar  )
\right)\\
   \hat \redpar &=
   \mathrm{Alg}(\seq u_\tr , \seq y_\tr, \projpar),\label{eq:inner}
\end{align}
\end{subequations}
where the inner-level optimization $\mathrm{Alg}(\cdot, \cdot, \projpar)$ is the reduced system identification algorithm:
\begin{equation}
\mathrm{Alg} (\seq u_\tr , \seq y_\tr, \projpar ) =
\arg \min_\redpar \loss(\seq y_\tr, \M(\seq u_\tr ; \, P_\projpar(\redpar))).
\end{equation}

Essentially,~\eqref{eq:inner} is a parametric version of~\eqref{eq:reduced_sysid} where the coefficients $\projpar$ of the lifting function $P_\projpar$ are free. In fact, $\projpar$ is optimized at the outer level~\eqref{eq:outer} to minimize the expected test set loss, according to the nested structure of the overall optimization problem~\eqref{eq:bilevel}.

\subsection{Encoder-decoder architecture}
\label{enc_dec_hypernet}
To circumvent the numerical intricacies related to the nested optimization problem formulation~\eqref{eq:bilevel}, we replace the inner optimization with an auxiliary neural network $E_{\encpar}$, which may be interpreted as a \emph{learned} black-box system identification algorithm:
\begin{subequations}
\label{eq:hypernet}
\begin{align}
\hat \projpar, \hat \encpar &= \arg \min_{\projpar, \encpar} \E_{p(D)}
\overbrace{\big [ \loss({\seq y}_\te, \hat {\seq y}_\te) \big ]}^{J(\projpar, \encpar)} \label{eq:hypernet_a}\\
\hat{\seq{y}}_\te  &= \M\left(
  \seq{u}_\te ; \,
  P_{\projpar} ( \hat \redpar )
\right) \label{eq:decoder}\\
  \hat \redpar &=
  E_{\encpar}(\seq u_\tr , \seq y_\tr ) \label{eq:encoder}.
\end{align}
\end{subequations}

\begin{figure}
\begin{center}
\includegraphics[width=.6\hsize]{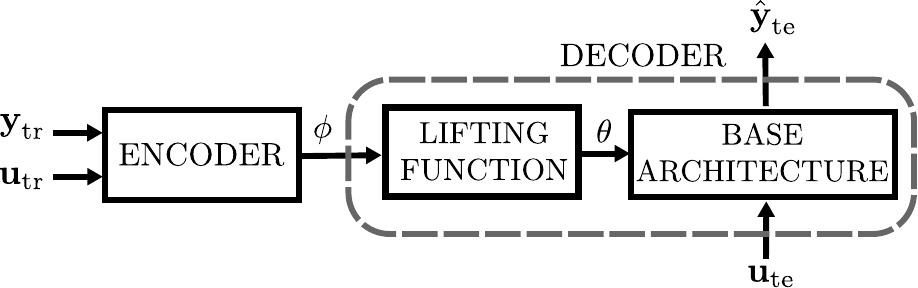}
\caption{The autoencoder architecture for reduced-complexity architecture learning.}
\label{fig:encdec_hypernet}
\end{center}
\end{figure}

The pair \eqref{eq:decoder}-\eqref{eq:encoder} may also be seen as an encoder-decoder architecture. The encoder $E_\encpar$ ingests the input-output pair $(\seq u_\tr , \seq y_\tr )$ and outputs reduced-complexity parameters $\hat \redpar$. The decoder, defined by the functional composition $\M(\cdot; P_{\hat \projpar}(\cdot))$, first lifts the reduced parameters  $\hat \redpar$ to the full-complexity dimension $n_\fullpar$, and then applies the \base\ with the lifted parameters to the input $\seq u_\te $ to predict the output $\hat {\seq y}_\te $. Figure~\ref{fig:encdec_hypernet} is a graphical representation of~\eqref{eq:decoder}-\eqref{eq:encoder}
focusing on this encoder-decoder interpretation.

To solve~\eqref{eq:hypernet}, we replace the intractable expectation over $p(D)$
by a Monte Carlo average. In other words, $J(\projpar, \encpar)$ in~\eqref{eq:hypernet} is approximated as a mean  over $b$ sampled datasets $D^\ii$ from the meta-dataset $\DD$ leading to:
\begin{equation}
    \label{eq:monte_carlo}
    \tilde J(\projpar, \encpar) = \frac{1}{b} \sum_{i=1}^b \loss\bigg(\seq y_\te^\ii, \M\Big(\seq u_\te^\ii; P_\projpar\big(E_\encpar(\seq u_\tr^\ii, \seq y_\tr^\ii)\big)\Big)\bigg)
\end{equation}
The loss $ \tilde J(\projpar, \encpar)$ is minimized with standard gradient-based optimization. In case the datasets $D^\ii$ are resampled at each iteration, $b$ plays the role of the batch size.

We emphasize that the main objective of solving \eqref{eq:hypernet} is to find the lifting function $P_{\hat \projpar}$ to be plugged into~\eqref{eq:rednet}, and, therefore, the manifold $M$ of dimension $n_\redpar$ that best reduces the complexity of identifying a model over the dataset distribution $p(D)$.
Once $P_{\hat \projpar}$ is learned, given a new dataset $D^* = (\seq{u}^*, \seq{y}^*)$, the user may prefer to solve the identification problem~\eqref{eq:reduced_sysid} explicitly, instead of relying on the learned $E_{\hat \encpar}$.
In fact, \eqref{eq:reduced_sysid} can be solved numerically using standard algorithms, typically with (local) convergence guarantees. These algorithms are arguably more trustworthy than the neural network $E_{\hat \encpar}$, which is a black-box from datasets to vectors of dimension $n_\redpar$ and only learns to approximate algorithmic behavior through the meta-learning procedure, thereby inheriting the usual generalization and representation challenges of learning-based methods.
Nonetheless, the learned algorithm $E_{\hat \encpar}$ may be used to initialize an iterative optimization algorithm for~\eqref{eq:reduced_sysid}, thereby speeding up the reduced-complexity model learning task.

\subsection{Specialization for neural state-space models}
We now specialize the methodology presented in the previous sections for datasets $D^\ii$ that are input/output time series generated by dynamical systems, and the \base\ is a state-space model:
\begin{equation}
\label{eq:generic_ss}
    x_{k+1} = f(x_k, u_k; \fullpar) \qquad
    y_{k} = g(x_k, u_k; \fullpar).
\end{equation}
In~\eqref{eq:generic_ss}, $f$ and $g$ are learnable mappings, such as feed-forward neural networks.
In this case, the output sequence $\seq y$, given an input sequence $\seq u$, also depends on an initial state $x_0$. Formally, we have: $\hat {\seq y} = \M(\seq u, x_0; \fullpar)$.
Thus, in applying the previously mentioned methodology, care has to be taken in the initialization of this state, notably for a correct computation of the loss in~\eqref{eq:hypernet_a}.

A few practical solutions have been introduced in recent system identification literature to tackle the initial state issue. Similarly to \cite{beintema2023deep,Bem25}, an additional neural network can be included to process  the first $n_{\rm init}$ samples of $(\seq u_\te , \seq y_\te )$ and provide an estimate $\hat x_{n_{\rm init}}$ of the state at time step $n_{\rm init}$. Starting from this state estimate, the loss~\eqref{eq:hypernet_a} may be computed, excluding the first $n_{\rm init}$ samples from the loss evaluation. We note that, in the case of synthetic data, the two datasets $(\seq u_\tr , \seq y_\tr )$, $(\seq u_\te , \seq y_\te )$ may be chosen to have the \emph{same} (random) initial state. In this case, the encoder network $E_\encpar$ may be designed to provide initial state estimates, together with the reduced-complexity model parameters, thereby simplifying the architecture.

Finally, an even simpler approach may be followed when the systems to be modeled are known to have a fading memory. In this case, the state can be initialized to an arbitrary value (such as zero), and the effect of the system's ``natural response'' can be removed from the loss by ignoring the contribution from the first $n_{\rm skip}$ samples, where $n_{\rm skip}$ is selected large enough. In the numerical examples presented in Section~\ref{sec:examples}, for simplicity, problem~\eqref{eq:hypernet} is instantiated relying on such a ``zero'' state initialization scheme. More advanced schemes may be integrated with a moderate effort.

\section{Probabilistic bound on the performance gap} \label{sec:prob_bounds}

In this section, we derive a probabilistic bound on the gap between training the full-complexity model \eqref{eq:fullnet}
and the reduced-complexity architecture \eqref{eq:rednet}.

For each dataset $D$ drawn from the dataset distribution $p(D)$,
let $(D_\tr, D_\te)$ be the  split of $D$
that partitions $D$ into a training part $D_\tr=(\seq u_\tr,\seq y_\tr)$ and a test part $D_\te=(\seq u_\te,\seq y_\te)$.
Note that $D_\tr$ and $D_\te$ are generated by the same underlying system (i.e., they share the same latent $\latent$ in \eqref{eq:dataset_distribution}). Given  $D_\tr$, let the trained parameters of the full- and reduced-complexity models be, respectively,
\begin{align}
\hat \fullpar(D_\tr) &= \arg \min_{\fullpar} \loss\!\left(\seq y_\tr,\M(\seq u_\tr;\fullpar)\right), \\
\hat \redpar(D_\tr) &= \arg \min_{\redpar} \loss\!\left(\seq y_\tr, \M(\seq u_\tr; P_{\hat\projpar}(\redpar))\right).
\end{align}
We define the fit index  over a set of length $N$ as
\begin{equation}
\label{eq:fit_index}
\mathrm{fit}(\seq{y}, \hat {\seq{y}}) = 100\cdot \left(1- \frac{\sqrt{\sum_{k=0}^{N-1} \left(\seq{y}_k -  \hat {\seq{y}}_k\right)^2} }
{\sqrt{\sum_{k=0}^{N-1} \left(\seq{y}_k -  {\overline{\seq{y}}}\right)^2}}\right) (\%),
\end{equation}
where ${\overline{\seq{y}}}$ is the sample average of the measured output $\seq y$.

Then, we define the \emph{performance gap} as the difference between the test fit index obtained using the full- and the reduced-complexity models, i.e.,
\begin{equation}
\label{eq:validation_gap}
\Psi(D) =
\mathrm{fit}\!\left(\!\seq y_\te,\M(\seq u_\te;\hat \fullpar(D_\tr))\!\right) - \mathrm{fit}\!\left(\!\seq y_\te,\M(\seq u_\te;P_{\hat\projpar}(\hat \redpar(D_\tr)))\!\right).
\end{equation}

The gap $\Psi(D)$ thus measures the  difference in the performance on a test set between full- and reduced-complexity models, identified based on the same training set $D_\tr$. If $\Psi(D) > 0$, the full-complexity model achieves a higher fit than the reduced-complexity  model, i.e., training in the full space yields better performance on $D_\te$. Conversely, if $\Psi(D) < 0$, the reduced-complexity model performs better on $D_\te$.

Theorem~\ref{thm:val-gap} defined below establishes a probabilistic upper bound for the performance gap $\Psi$. Such a bound is constructed using $m$ \emph{additional} independent and identically distributed (i.i.d.) datasets $\{D^{(i)}\}_{i=1}^m$
extracted from $p(D)$, independent of the datasets used to learn $\projpar$ and $\encpar$ in the previous meta-learning phase.

\begin{theorem}[Bound on the performance gap]
\label{thm:val-gap}
\ \\
Let $\{D^{(i)}\}_{i=1}^m$ be i.i.d.\ draws from $p(D)$, independent of the datasets previously used in the meta-learning phase.
For each $i$ let $\bigl(D_\tr^{(i)}, D_\te^{(i)}\bigr)$ be the corresponding training/test split.
Define $\Psi^{(i)} = \Psi\!\bigl(D^{(i)}\bigr)$
and let $\Psi_{(1)} \le \cdots \le \Psi_{(m)}$ be the order statistics of the samples $\{\Psi^{(i)}\}_{i=1}^m$, i.e., the sequence of    $\{\Psi^{(i)}\}_{i=1}^m$ arranged in non-decreasing order.

For any $\alpha,\delta\in(0,1)$, set
$\varepsilon_m = \sqrt{\frac{\log(2/\delta)}{2m}}$ and
$k^\star = \left\lceil m\bigl(1-\alpha+\varepsilon_m\bigr)\right\rceil.$
Suppose $m$ is large enough so that $k^\star \le m$ (equivalently, $\alpha \ge \varepsilon_m$).
Let $D^{\mathrm{new}}\sim p(D)$ be a new  draw independent of  $\{D^{(i)}\}_{i=1}^m$, with training/test split
$\bigl(D_\tr^{\mathrm{new}}, D_\te^{\mathrm{new}}\bigr)$, and define
\begin{equation}
\Psi^{\mathrm{new}} = \Psi\!\bigl(D^{\mathrm{new}}\bigr).
\end{equation}
Then, with probability at least $1-\delta$ over the draw of the training/test pairs $\{D^{(i)}\}_{i=1}^m$, it holds:
\begin{align}
\mathbb{P}\bigl(\Psi^{\mathrm{new}} \le \Psi_{(k^\star)}  \bigr)
\;\ge\; 1-\alpha,
\label{eq:main-val-gap}
\end{align}
where $\mathbb{P}$ denotes the probability measure induced by the sampling distribution $p(D)$.
\end{theorem}

\paragraph*{Proof.}

The proof builds upon the same rationale   adopted in \cite[Section~4.2]{BuMe2025} in the context of amortized optimization
to derive  bounds on the gap between solving an optimization problem
and its reduced-dimensional counterpart learned from data.
A   proof is provided in the following for self-consistency  of the paper.

First, since the datasets $\{D^{(i)}\}_{i=1}^m$ are  i.i.d. draws from $p(D)$ and the mapping $D \mapsto \Psi(D)$ is deterministic (given the training/test procedure), the  performance gaps $\{\Psi^{(i)}\}_{i=1}^m$ are i.i.d.\ random variables.

Let $\CFD$ be the \emph{cumulative distribution function} (CDF) of $\Psi(D)$ and let $\hat{\CFD}_m$ be its empirical CDF defined as:
\begin{align} \label{eqn:CDFemp}
\hat{\CFD}_m(t) \;=\; \frac{1}{m}\sum_{i=1}^m \mathbf{1}\bigl(\Psi^{(i)} \le t \bigr),
\qquad t \in \mathbb{R},
\end{align}
with $\mathbf{1}$ being the indicator function.

By the Dvoretzky–Kiefer–Wolfowitz–Massart inequality~\cite{massart1990tight}, the following condition holds with probability at least $1-\delta$ over the  sets $\{D^{(i)}\}_{i=1}^m$:
\begin{align}
\sup_{t} \bigl| \hat{\CFD}_m(t) - \CFD(t) \bigr| \;\le\; \varepsilon_m.
\label{eq:DKW}
\end{align}
  For $\alpha \in (0,1)$, let $Q_{1-\alpha} = \inf\{t : \CFD(t) \ge 1-\alpha\}$ be the $(1-\alpha)$-quantile of the distribution of $\Psi(D)$.
Let us choose  $k^\star = \Bigl\lceil m\bigl(1-\alpha + \varepsilon_m \bigr)\Bigr\rceil \le m.$
Then,
\begin{align}
\hat{\CFD}_m\bigl(\Psi_{(k^\star)}\bigr) \;\ge\; \frac{k^\star}{m} \;\ge\; 1-\alpha+\varepsilon_m,
\label{eq:hatF}
\end{align}
where the first inequality follows from the definition of the empirical CDF $\hat{G}_m$ in \eqref{eqn:CDFemp}, evaluated at the $k^\star$-th order statistic $\Psi_{(k^\star)}$, and the second inequality follows from the definition of $k^\star$.

Combining \eqref{eq:DKW} and \eqref{eq:hatF}, with probability at least $1-\delta$ we obtain
\begin{align}
\CFD\bigl(\Psi_{(k^\star)}\bigr)
\;\ge\; \hat{\CFD}_m\bigl(\Psi_{(k^\star)}\bigr) - \varepsilon_m
\;\ge\; 1-\alpha,
\end{align}
so that $Q_{1-\alpha} \le \Psi_{(k^\star)}$ by the definition of $Q_{1-\alpha}$.

Finally, for a new draw $D^{\mathrm{new}}\sim p(D)$, we have
\begin{align}
\mathbb{P}\bigl(\Psi^{\mathrm{new}} \le Q_{1-\alpha}\bigr) = 1-\alpha
\end{align}
by definition of $Q_{1-\alpha}$. Since we proved that  $Q_{1-\alpha}\le \Psi_{(k^\star)}$,  then
\begin{align}
\mathbb{P}\bigl(\Psi^{\mathrm{new}} \le \Psi_{(k^\star)}\bigr)
\;\ge\; \mathbb{P}\bigl(\Psi^{\mathrm{new}} \le Q_{1-\alpha}\bigr)
= 1-\alpha,
\end{align}
which proves \eqref{eq:main-val-gap}.
\hfill$\square$

Note that, although Theorem \ref{thm:val-gap} is stated for the  performance gap defined in \eqref{eq:validation_gap}, the same argument applies to other performance metrics, for instance, the loss $\loss$ of a new dataset  $D^{\mathrm{new}}$.

\begin{remark}
To ensure that the index $k^\star$ is well-defined, one needs $m \ge k^\star$.
This yields the requirement~$\displaystyle
m \ge \frac{\log(2/\delta)}{2\alpha^2}$,
showing that the minimum number of datasets $m$ needed to obtain certification bounds scales logarithmically with $1/\delta$ and quadratically with $1/\alpha$. Nevertheless, it is worth noting that the $m$ datasets used to compute the bound on the performance gap can often be generated in simulation, by sampling i.i.d.  synthetic datasets from the same dataset distribution $p(D)$ employed in the meta-training phase (Section~\ref{sec:reduced_order}).
\end{remark}

\begin{remark}
It is worth clarifying the different roles played by the datasets $D$ used in the meta-learning phase (Section~\ref{sec:reduced_order}) and in the certification phase (Section~\ref{sec:prob_bounds}).
In the \emph{meta-training} phase, at each optimization step of the meta-objective~\eqref{eq:monte_carlo}, we sample $b$ datasets from $p(D)$ to approximate the expectation in \eqref{eq:hypernet_a} by the empirical average in~\eqref{eq:monte_carlo}. This procedure is used to learn the shared representation, namely the encoder $E_\encpar$ and the lifting function $P_\projpar$.
In the \emph{certification} phase, after $(\projpar,\encpar)$ have been learned and fixed, we draw an independent collection of $m$ previously unseen i.i.d. datasets from $p(D)$ to compute the performance gap $\Psi(D)$ and derive  probabilistic guarantees (Theorem~\ref{thm:val-gap}).
\end{remark}

\section{Numerical illustration}
\label{sec:examples}
We apply our approach to the Bouc--Wen system identification benchmark~\cite{noel2016hysteretic}.
The software was developed in Python using the JAX automatic differentiation library and it is available at the repository \url{https://github.com/forgi86/sysid-neural-manifold}. All computations were performed on a server equipped with an AMD EPYC 9754 CPU and an Nvidia 4090 GPU.

\subsection{The Bouc--Wen benchmark}
The Bouc--Wen benchmark entails the vibrations of a 1-DoF hysteretic mechanical system. The system dynamics are described in state-space form by:
\begin{subequations}
\label{eq:boucwen}
\begin{align}
\begin{bmatrix}
\dot{p}(t) \\
\dot{v}(t) \\
\dot{z}(t)
\end{bmatrix}
&=
\begin{bmatrix}
v(t) \\
\frac{1}{m_L} \left (u(t) -k_L p(t) - c_L v(t) - z(t)  \right) \\
\alpha v(t) - \beta(\gamma |v(t)| |z(t)|^{\nu -1} + \delta v(t) |z(t)|^\nu)
\end{bmatrix}\\
y(t) &= p(t),
\end{align}
\end{subequations}
where $p(t)$~(m) is the measured output position; $v(t)$~(m/s) is the unmeasured velocity; $z(t)$~(N) is an unmeasured hysteretic force term; and $u(t)$~(N) is the known input force. The nominal values of the model coefficients are reported in the first row of Table~\ref{tab:bw_params}.
\begin{table}[!ht]
\setlength{\tabcolsep}{4pt}
\centering
\caption{Coefficients of the Bouc--Wen system. Nominal values (first row) and minimum/maximum values in the dataset distribution (second/third row).}
\label{tab:bw_params}
\small
\begin{tabular}{c||cccccccc}
\toprule
 & $m_L$ & $c_L$ & $k_L$ & $\alpha$ & $\beta$ & $\gamma$ & $\delta$ &$\nu$\\
\midrule
nom & 2 & 10 &$5.0\times 10^4$ & $5.0\times 10^4$ & 1000 & 0.8 & -1.1 & 1\\
min & 1 & 5 & $2.5\times 10^4$ & $2.5\times 10^4$ & $500$ & 0.5 & -1.5 &1\\
max & 3 & 15 & $7.5\times 10^4$ & $7.5\times 10^4$ & 4500 & 0.9 & -0.5 &1\\
\bottomrule
\end{tabular}
\end{table}

The benchmark provides a test dataset where the input $u(t)$ is a random-phase multisine of length $8192$ sampled at frequency $f_s=750$~Hz  that excites the frequency range $[5, 150]$~Hz and has a root mean square of $50$~N. System~\eqref{eq:boucwen} is simulated with
its nominal coefficients and the test output $y(t) = p(t)$ is provided noise-free.
A ``default'' training dataset is also provided,  comprising 40960 samples generated by exciting the system with a multisine input signal with the same characteristics as the test input. In the training dataset, the output is corrupted by an additive Gaussian noise with bandwidth $[0, 375]$~Hz and amplitude $8 \cdot 10^{-3}$. Moreover, the benchmark provides simulation code and specifies that the user is free to choose different input signals to excite the system and generate arbitrary training datasets.

In this paper, along the lines of \cite{chakrabarty2025meta}, we extend the original scope of the benchmark~\cite{noel2016hysteretic} and use the simulator to generate additional datasets with perturbed coefficient values, aiming at finding a reduced-complexity architecture that provides high performance over the entire \emph{class} of Bouc--Wen systems with parameters comprised in the min-max range reported in Table~\ref{tab:bw_params}. Importantly, meta-learning of the reduced-complexity architecture is performed exclusively on these additional datasets. The training and test datasets provided by the benchmark developers and generated using the nominal coefficients are only used at a later stage to \emph{assess} the performance of the meta-learned architecture. This strategy prevents the risk of \emph{meta-overfitting} to a specific Bouc--Wen configuration. In meta-learning terminology, the new datasets generated by the authors of this work constitute the \emph{meta-training} set, while the benchmark-provided datasets serve as the \emph{meta-test} set.

\subsection{Base architecture}
As \base,\ we consider the neural state-space:
\begin{subequations}
\label{eq:ss_boucwen}
\begin{align}
    x_{k+1} &= A x_k + B u_k + N_f(x_k, u_k; W_f) \\
    y_{k} &= C x_k +  N_g(x_k; W_g),
\end{align}
\end{subequations}
where $n_u=1$, $n_y=1$, $n_x=3$, $N_f$, and $N_g$ are feed-forward neural networks with one hidden layer and $16$ hidden units with weights $W_f$ and $W_g$, respectively, and \texttt{tanh} activation function; and $A$, $B$, and $C$ are matrices of proper size. The \base\ parameters are thus $\fullpar = \mathrm{vec}(W_f,\ W_g,\ A,\ B,\ C) \in \R^{n_\fullpar}$, with $n_\fullpar=244$.

\subsection{Metrics and baselines}
\begin{figure}
    \centering
    \includegraphics[width=.8\hsize]{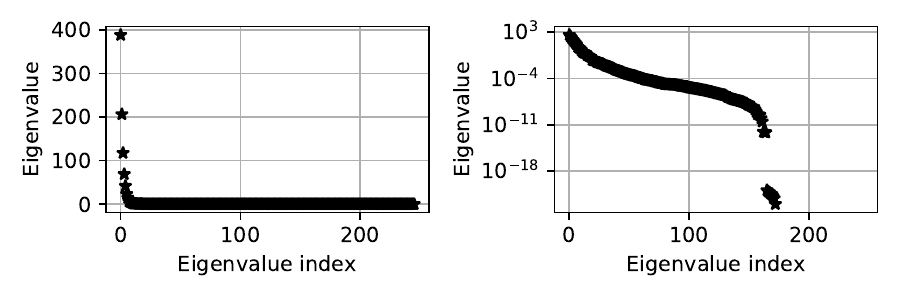}
    \caption{Eigenvalues of the Hessian of the full-complexity training loss: eigenvalues on a linear (left panel) and logarithmic (right panel) scale.}
    \label{fig:boucwen_hessian}
\end{figure}

We evaluate performance in terms of the $\mathrm{fit}$ index~\eqref{eq:fit_index}.
Fitting the model on the default $40960$-length training dataset, \cite{schoukens2021improved} reported the state-of-the-art result on this benchmark ($\mathrm{fit}=98.9~\%$) on the test set. We could match their results by minimizing, in line with the full-complexity training setting~\eqref{eq:full_sysid}, the MSE loss of the \base\ \eqref{eq:ss_boucwen} initialized with random parameters over 40000 iterations of \texttt{AdamW} with learning rate $10^{-3}$ followed by 10000 iterations of \texttt{BFGS}. The training time was~1.4 hours.
The eigenvalues of the Hessian of the full-complexity training loss at the solution of the optimization problem are visualized in Figure~\ref{fig:boucwen_hessian}.
Interestingly, many of those eigenvalues are close to zero, which suggests that the model structure is over-parameterized and further motivates the pursuit of a lower-dimensional parameterization.

We also train as a baseline a third-order linear model that provides $\mathrm{fit}=77.2~\%$, underlining the relevance of non-linearities in this benchmark.

\subsection{Full-complexity model training}
\label{sec:full_order_training}

In this section, we assess the performance of full-complexity model learning in the low-data regime. To this aim, we fit full-complexity models according to~\eqref{eq:full_sysid} over sequences of length
$L=$ $100$, $200$, $400$, $500$, $600$, $800$, $1000$, $2000$, $3000$, $4000$, $5000$.
 As in the previous section, the base architecture is trained from scratch starting from a random initialization, thus without leveraging any prior information from the dataset distribution $p(D)$.
For each length $L$, we execute a Monte Carlo study involving $n_{\rm mc}=100$ subsequences of length $L$ extracted from the official $40960$-length dataset, training distinct models on each sequence.

\begin{figure}
    \centering
    \includegraphics[width=\textwidth]{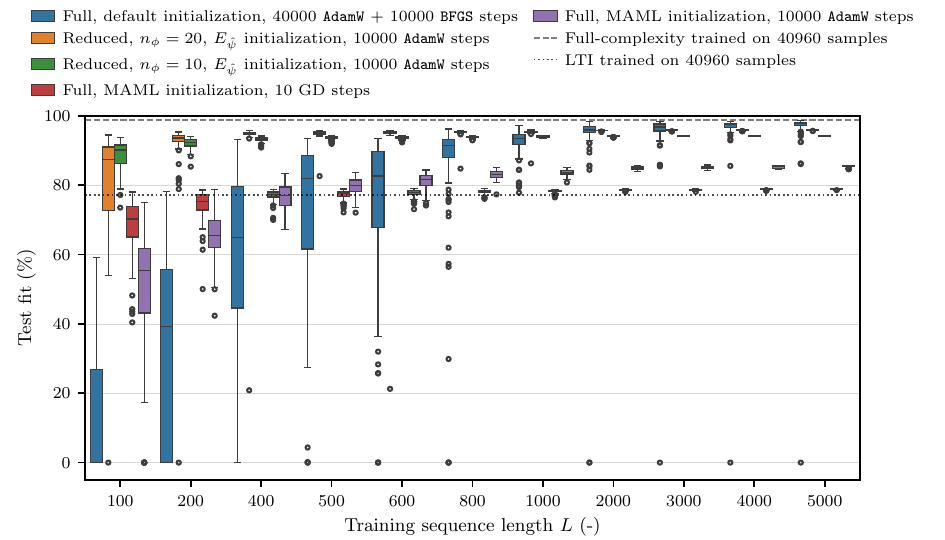}
    \caption{Full- and reduced-complexity models: test $\mathrm{fit}$ vs. training sequence length $L$.}
    \label{fig:boucwen_boxplot}
\end{figure}

For optimization, we use the same \texttt{AdamW} + \texttt{BFGS} approach with the same settings detailed in the previous section.
As shown in Figure~\ref{fig:boucwen_boxplot}, full-complexity model learning proves highly inefficient in the low-data regime. In particular, for sequence length $L \leq 400$, the median $\mathrm{fit}$ is below the linear baseline.
The median performance consistently improves for increasing sequence length, with average \textrm{fit} above $97~\%$ for $L\geq 3000$ samples.
However, even for $L=2000, 3000, 4000$, and $5000$, a small number of runs (2, 1, 1, and 1, respectively) fail due to numerical issues in the \texttt{BFGS} solver. While these issues might be mitigated with a careful tuning of the solver settings, they highlight the challenges of full-complexity model training.

\begin{figure}
\centering
    \includegraphics[width=.8\hsize, trim=10 10 10 10]{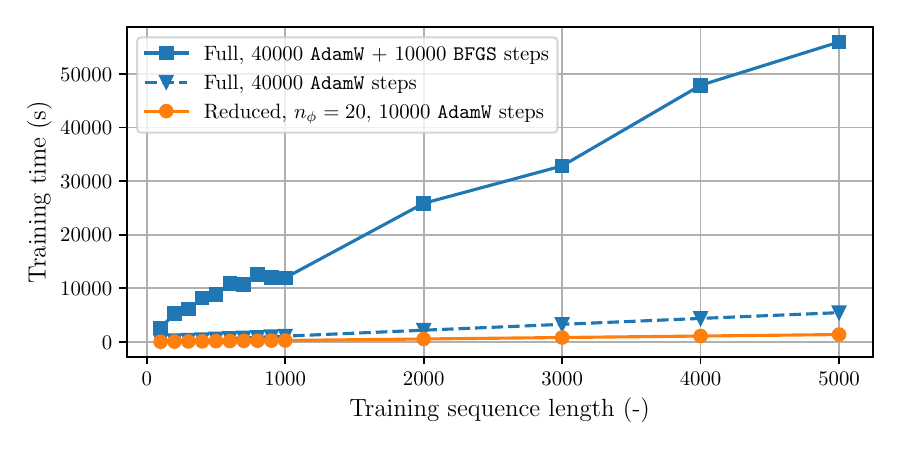}
    \caption{Training time vs. training sequence length $L$.}
    \label{fig:training_time}
\end{figure}

Figure~\ref{fig:training_time} illustrates the time required to train in parallel the 100 models vs. the sequence length $L$. We report both the run time for the $40000$ \texttt{AdamW} iterations and the total training time, which includes the $10000$ \texttt{BFGS} iterations. The results show that \texttt{BFGS} dominates the overall training time.

\subsection{Meta-learning the reduced-complexity architecture}
\label{sec:meta_training}
To enable higher performance in the low-data regime, we learn a manifold with $n_\redpar=20$ parameters, embedded within the full-complexity architecture~\eqref{eq:ss_boucwen} characterized by~$n_\fullpar=244$ parameters, by applying the methodology described in Section~\ref{sec:reduced_order}.
To this aim, we introduce the parameterized lifting network $P_\projpar : \R^{n_\redpar} \mapsto \R^{n_\fullpar} = V \redpar + \fullpar_{\rm bias}$ with $V \in \R^{n_\fullpar \times n_\redpar}, \fullpar_{\rm bias} \in \R^{n_\fullpar}$,  which expands the reduced-complexity dimension $n_\redpar$ to the full-complexity $n_\fullpar$. Its tunable parameters are collected in the vector $\projpar = \mathrm{vec}(V, \fullpar_{\rm bias}) \in \R^{n_\projpar}$, with $n_\projpar = (n_\redpar + 1) n_\fullpar = 5124$.

 To learn the optimal lifting network parameters $\hat \projpar$, we adopt the encoder-decoder methodology and solve the optimization
problem~\eqref{eq:hypernet}, using the Monte Carlo approximation $\tilde J(\projpar, \encpar)$ in~\eqref{eq:monte_carlo}
of the loss $J(\projpar, \encpar)$.
The meta-dataset $\DD$ is obtained by randomly sampling Bouc--Wen systems with coefficients uniformly distributed within the min-max ranges in Table~\ref{tab:bw_params}, and feeding as input random multisine sequences of length $N=2000$ with the same properties as the default training input.

The encoder $E_\encpar$ is a sequence-to-vector architecture obtained by combining a bi-directional Gated Recurrent Unit (GRU) with a feed-forward neural network. The GRU is configured with $n_u+n_y=2$ input channels and a single hidden layer with $n_h=128$ hidden units.
It processes the sequences $(\seq u_\tr , \seq y_\tr )$ and produces a sequence of activations  $\seq h_{0:N-1}$, $\seq{h}_i \in \R^{n_h}$. This sequence is reduced to a single vector $h_m$ through \emph{average pooling}: $h_m = \frac{1}{N} \sum_i{\seq {h}_i}$ and further processed by the feed-forward neural network configured with
$n_h=128$ inputs, a hidden layer with 128 units, and an output layer with
$n_\redpar = 20$ units, and \texttt{tanh} activation functions. Overall, the network $E_\encpar$ has $n_\encpar=136340$ parameters corresponding to the GRU and the feed-forward neural network weights. It takes as input the sequences $(\seq u_\tr , \seq y_\tr )$ and produces as output a single vector of size $n_\redpar$.

For gradient-based minimization of the loss~\eqref{eq:monte_carlo}, the \texttt{Adam} algorithm is applied with a batch size $b=128$ and initial learning rate of $2 \cdot 10^{-4}$, decreased to $2 \cdot 10^{-5}$ with cosine scheduling over $200000$ iterations. The datasets $D^{(i)}$ are resampled at each meta-optimization iteration. Thus, overall, 256 million trajectories are used for meta training.
The optimization takes approximately 25.3 hours. As a result, we obtain the reduced-complexity architecture
\eqref{eq:rednet} and, as a byproduct, the encoder network $E_{\hat \encpar}$ which can be used to initialize the reduced-complexity model learning~\eqref{eq:reduced_sysid}.

\subsection{Reduced-complexity model training}
\label{sec:reduced_order_training}
We repeat the Monte Carlo study of Section~\ref{sec:full_order_training} involving $n_{\rm mc}=100$ datasets per sequence length using the reduced-complexity architecture meta-learned in Section~\ref{sec:meta_training}, instead of the \base.

In all the runs, the reduced-complexity MSE loss is minimized according to~\eqref{eq:reduced_sysid}. Specifically, minimization is performed over $10000$ iterations of \texttt{AdamW} with learning rate $10^{-3}$, using the encoder network $E_{\hat \encpar}$ to obtain an initial guess of $\redpar$. Results in Figure~\ref{fig:boucwen_boxplot} show that the reduced-complexity training is significantly more effective than the full-complexity one for sequences of length up to $2000$ samples. Very good results are obtained with just $500$ samples (median fit of $95.2\%$). For training sequences longer than $2000$ samples, full-complexity training generally yields superior results, given the higher flexibility of the model structure.

We remark that none of the reduced-complexity training runs exhibited numerical difficulties, and the obtained performance is largely insensitive to the choice of the optimizer and its settings. This underscores the stability and reliability of the reduced-complexity model learning problem.

The run time of reduced-complexity training with $10000$ \texttt{AdamW} iterations is shown in Figure~\ref{fig:training_time} and amounts to approximately $25$\% of the $40000$ \texttt{AdamW} iterations executed in the full-complexity case. This highlights that the cost of a \emph{single} optimization iteration remains nearly identical for both cases, as evaluating the lifting function $P$ is negligible compared to computing the training loss. Indeed, $P$ is a lightweight static function evaluated only once per iteration, while the neural state-space~\eqref{eq:fullnet} is executed iteratively over the $L$ time steps.

Due to the lower problem dimensionality and to the near-optimal parameter initialization
using $E_{\hat \encpar}$, $10000$ \texttt{AdamW} iterations are sufficient for reduced-complexity optimization.
It is worth reminding  that this efficiency gain is achieved after the expensive meta-training phase, which however has to be executed only once for the whole class of identification problems described by the probability distribution $p(D)$.

To analyze the sensitivity of the methodology to the choice of the manifold dimensionality, we repeat the reduced-complexity architecture meta-learning and the Monte Carlo study for $n_\redpar=10$. Results in Figure~\ref{fig:boucwen_boxplot} highlight that this choice leads to the best results for the very short sequences of length $L=100$, while it leads to worse performance against $n_\redpar=20$ for all other lengths.

\subsection{Empirical evaluation of the performance gap bound}
In this section, we numerically assess the relevance
of  Theorem~\ref{thm:val-gap}, namely that the bound~\eqref{eq:main-val-gap} on the performance gap between the full- and reduced-complexity training schemes holds with a (high) probability of at least \(1-\delta\). To this end, we set \(\delta=0.01\), \(\alpha=0.05\), and perform $m=1900$  full and reduced trainings on
$m$ datasets of length $L=500$ and $L=3000$ extracted from $p(D)$.
For each run, we record the performance gap statistic  $\Psi$ defined in~\eqref{eq:validation_gap}. The results are sorted in non-decreasing order and denoted by \(\{\Psi_{(i)}\}_{i=1}^m\).

\begin{figure}
    \centering
    \includegraphics[width=.8\linewidth]{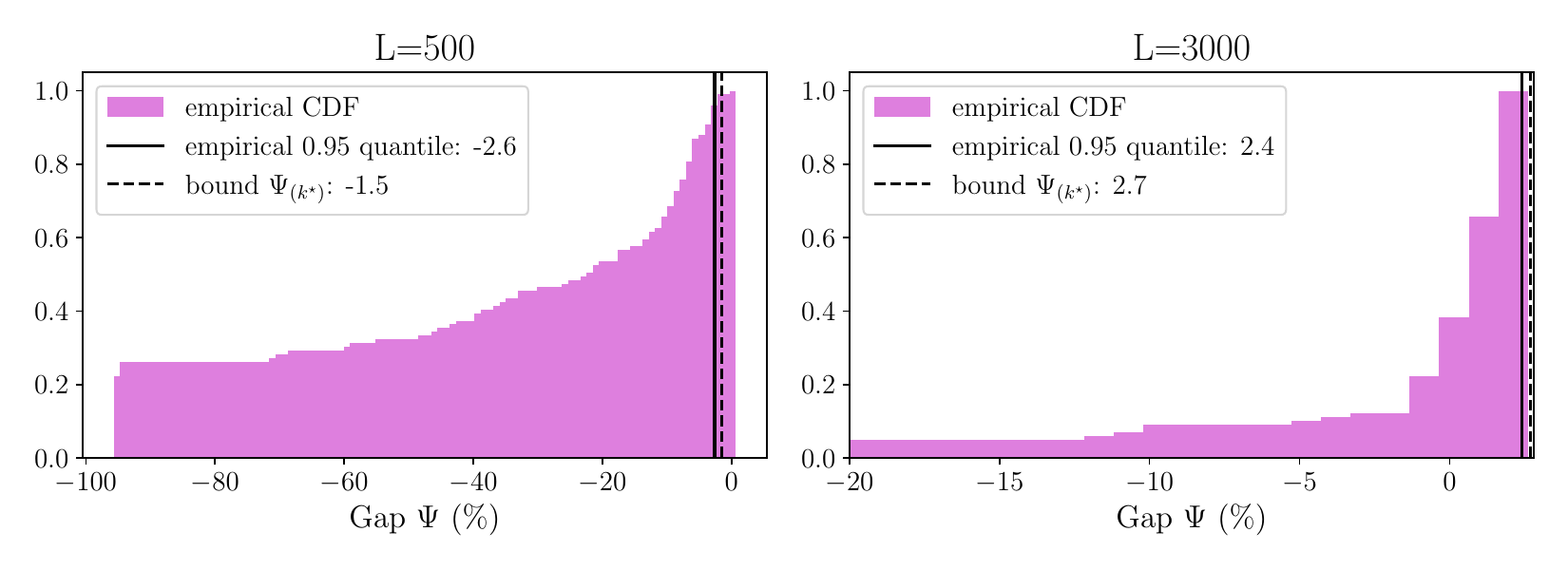}
    \caption{Performance gap ($\mathrm{fit}$ index difference) between full and reduced models: empirical cumulative distribution function (pink), empirical 0.95 quantile (solid black line) and theoretical upper bound
    $\Psi_{(k^\star)}$ on such quantile. Results for $L=500$ (left panel) and $L=3000$ (right panel).}
    \label{fig:performance_gap}
\end{figure}

Given the chosen values of $\alpha$, $\delta$, and $m$, we obtain $\varepsilon_m = 0.037$ and $k^\star = 1876$. The element $\Psi_{(k^\star)}$
in the ordered set \(\{\Psi_{(i)}\}_{i=1}^m\) then serves as a probabilistic upper bound on the performance gap, as established by~\eqref{eq:main-val-gap}. Figure~\ref{fig:performance_gap} illustrates this bound for both sequence lengths, together with the
empirical cumulative distribution function (CDF) of $\Psi$ obtained from an additional 100 Monte Carlo simulations. We observe that the empirical
$(1-\alpha)$-quantile of $\Psi$ in these additional simulations is indeed bounded by $\Psi_{(k^\star)}$, thereby illustrating the practical relevance of the theoretical result. For $L=500$, the bound $\Psi_{(k^\star)} = -1.5\%$ is negative, which confirms that reduced-complexity models achieve {better} performance on short sequences. In contrast, for $L=3000$, the bound $\Psi_{(k^\star)} = 2.7\%$ is positive, and upper bounds (with probability at least $1-\delta$) the potential performance loss of reduced-complexity model training.

\subsection{Comparison with MAML}
We perform a comparison with the MAML meta-learning approach. In the meta-learning phase, MAML solves an explicit bi-level optimization, similar to \eqref{eq:bilevel}, to find an initial full-complexity parameter $\fullpar_{\rm maml} \in \mathbb{R}^{n_\fullpar}$ for subsequent gradient-based optimization. In its inner level, MAML executes plain gradient descent initialized from $\fullpar_{\rm maml}$.
In our implementation, MAML is configured with 10 inner gradient-descent steps at an inner learning rate of $0.1$. Note that these are typical settings for this algorithm, and that {the number of inner gradient-descent steps is limited by computational and memory requirements.} The outer MAML optimization is solved with the same settings used for reduced-complexity architecture meta learning, i.e. with $200000$ \texttt{Adam} iterations and cosine learning rate scheduling. The MAML meta-learning optimization takes $3.3$ days.

The meta-learned parameter $\fullpar_{\rm maml}$ is then used as an initializer in the Monte Carlo study. Specifically, full-complexity trainings~\eqref{eq:full_sysid} starting from $\fullpar_{\rm maml}$ are executed both
with 10 steps of plain gradient descent (matching the MAML inner-level meta-training, that is however limited by computational reasons) and with 100000 iterations of \texttt{AdamW} with learning rate $1 \cdot 10^{-3}$ (matching the reduced-complexity  model trainings). All results are reported in Figure~\ref{fig:boucwen_boxplot}. We observe that MAML initialization with 10 GD steps is rather effective for the short sequences (though not as effective as training in the low-dimensional meta-learned manifold), but its performance barely surpasses the linear baseline for longer ones. Conversely, training
with $10000$ \texttt{AdamW} steps reaches a higher performance for the longer sequences, but it is less effective on the shorter ones. Overall, the reduced-complexity model training approach presented in this work dominates the MAML implementation.

The lower performance of MAML may be partly due to computational limits on the number of inner-optimization steps and the fact that its hyper-parameters were not extensively tuned for this benchmark. Moreover, the considered setting of an infinite meta-dataset likely mitigates the weaker inductive bias of our approach based on auxiliary black-box maps, which in turn may mask one of its drawbacks and contribute to its observed advantage over MAML.

\section{Conclusions}
\label{sec:conclusions}
This paper introduced a meta-learning framework for system identification that discovers a low-dimensional manifold within the parameter space of an over-parameterized neural architecture. Our results on the Bouc--Wen benchmark highlight the potential of this approach, particularly in extremely data-scarce scenarios.

Several extensions of this framework can be explored. While our methodology was introduced in the context of a black-box base architecture, it could be extended to models, constraints, and fitting criteria informed by physical knowledge.  Additionally, integrating variational learning techniques into the meta-learning process could enable the discovery of not only a deterministic low-dimensional representation but also a prior distribution over this space.

Finally, it would be interesting to study meta-training in the regime of small \emph{meta}-datasets, that is, when only a limited number of trajectories from a class of systems is available. In this context, statistical efficiency at the meta-learning procedure becomes critical, and a probabilistic formulation may play a key role by regularizing it through a suitable prior over the low-dimensional representation, thereby improving sample efficiency.

\section*{Acknowledgments}
This work was supported as a part of NCCR Automation, a National Centre of Competence in Research, funded by the Swiss National Science Foundation (grant number 51NF40\_225155). This research was supported by the European Research Council (ERC), Advanced Research Grant COMPACT (Grant Agreement No. 101141351).

\bibliographystyle{abbrv}
\bibliography{biblio}

\end{document}